\documentclass[letterpaper, 10 pt, conference]{URL-ieeeconf}
\IEEEoverridecommandlockouts    
\usepackage[vlined,ruled,linesnumbered]{algorithm2e}
\usepackage{graphics}
\usepackage{times}              
\usepackage{amsmath}
\usepackage{amssymb}
\usepackage{bm}
\usepackage[noend]{algpseudocode}
\usepackage{booktabs}
\usepackage{color}
\usepackage{subcaption}
\usepackage{rotating}
\usepackage{cite}
\usepackage{xstring}
\usepackage{hhline}
\usepackage{siunitx}
\usepackage{tikz}
\usepackage{soul}
\usepackage{tcolorbox}
\usepackage{textcomp}
\usepackage{array,multirow,graphicx}
\usepackage{stmaryrd}
\usepackage{hyperref}
\definecolor{instructioncolor}{rgb}{.5,.5,.5}

\definecolor{lightgray}{RGB}{200, 200, 200}
\sethlcolor{lightgray}

\def\secref#1{Section~\ref{#1}}
\def\figref#1{Fig.~\ref{#1}}
\def\tabref#1{Table~\ref{#1}}
\def\eqref#1{(\ref{#1})}

\def\vsfig{\vspace{-0.3cm}}

\newcommand\corresp{\mathcal{A}}

\newcommand\srccloud{\mathcal{P}}
\newcommand\srcpoint{\va}

\newcommand\tgtcloud{\mathcal{Q}}
\newcommand\tgtpoint{\vb}

\newcommand\srcidx{i}
\newcommand\tgtidx{j}
\newcommand{\kth}{k}

\newcommand{\rom}[1]{\uppercase\expandafter{\romannumeral #1\relax}}

\makeatletter
\usepackage{xspace}
\DeclareRobustCommand\onedot{\futurelet\@let@token\@onedot}
\def\@onedot{\ifx\@let@token.\else.\null\fi\xspace}
\makeatother

\newcolumntype{L}[1]{>{\raggedright\let\newline\\\arraybackslash\hspace{0pt}}m{#1}}
\newcolumntype{C}[1]{>{\centering\let\newline\\\arraybackslash\hspace{0pt}}m{#1}}
\newcolumntype{R}[1]{>{\raggedleft\let\newline\\\arraybackslash\hspace{0pt}}m{#1}}

\usepackage{pifont}
\newcommand{\bdmath}{\begin{dmath}}
\newcommand{\edmath}{\end{dmath}}
\newcommand{\beq}{\begin{equation}}
\newcommand{\eeq}{\end{equation}}
\newcommand{\bdm}{\begin{displaymath}}
\newcommand{\edm}{\end{displaymath}}
\newcommand{\bea}{\begin{eqnarray}}
\newcommand{\eea}{\end{eqnarray}}
\newcommand{\beal}{\beq \begin{array}{ll}}
\newcommand{\eeal}{\end{array} \eeq}
\newcommand{\beas}{\begin{eqnarray*}}
\newcommand{\eeas}{\end{eqnarray*}}
\newcommand{\ba}{\begin{array}}
\newcommand{\ea}{\end{array}}
\newcommand{\bit}{\begin{itemize}}
\newcommand{\eit}{\end{itemize}}
\newcommand{\ben}{\begin{enumerate}}
\newcommand{\een}{\end{enumerate}}

\newcommand{\calS}{{\cal S}}

\newcommand{\setal}{~\emph{et~al.}\xspace}
\newcommand{\eg}{\emph{e.g.,}\xspace}
\newcommand{\ie}{\emph{i.e.,}\xspace}

\def\etalcite#1{\setal~\cite{#1}}

\newcommand{\M}[1]{{\bm #1}} 
\renewcommand{\boldsymbol}[1]{{\bm #1}}

\newcommand{\hide}[1]{}

\newcommand{\hiddenText}{{\color{gray} hidden text.}}
\newcommand{\hideWithText}[1]{\hiddenText}

\DeclareMathOperator*{\argmin}{arg\,min}

\newcommand{\twonorm}[1]{\|#1\|_{2}}

\newcommand{\Real}[1]{ { {\mathbb R}^{#1} } }

\newcommand{\SOthree}{\ensuremath{\mathrm{SO}(3)}\xspace}

\newcommand{\MR}{\M{R}}

\newcommand{\va}{\boldsymbol{a}}
 
\newcommand{\vb}{\boldsymbol{b}}

\newcommand{\vt}{\boldsymbol{t}}

\newcommand{\vepsilon}{\boldsymbol{\epsilon}}

\newcommand{\blue}[1]{{\color{blue}#1}}

\newcommand{\linkToPdf}[1]{\href{#1}{\blue{(pdf)}}}
\newcommand{\linkToPpt}[1]{\href{#1}{\blue{(ppt)}}}
\newcommand{\linkToCode}[1]{\href{#1}{\blue{(code)}}}
\newcommand{\linkToWeb}[1]{\href{#1}{\blue{(web)}}}
\newcommand{\linkToVideo}[1]{\href{#1}{\blue{(video)}}}
\newcommand{\linkToMedia}[1]{\href{#1}{\blue{(media)}}}
\newcommand{\award}[1]{\xspace} 

\newcommand{\omitted}[1]{}

\newcommand{\bmat}{\left[ \begin{array}}
\newcommand{\emat}{\end{array}\right]}

\newcommand{\subMeas}[1]{\calS} %

\title{\LARGE \bf KISS-Matcher: Fast and Robust Point Cloud Registration Revisited}

\author{Hyungtae Lim$^{1}$, Daebeom Kim$^{2}$, Gunhee Shin$^{2}$, Jingnan Shi$^{1}$, Ignacio Vizzo$^{3}$, \\ Hyun Myung$^{2\dagger}$, Jaesik Park$^{4\dagger}$, and Luca Carlone$^{1\dagger}$
  \thanks{$^\dagger$This work was co-advised. \hfill \break
  \indent $^{1}$Laboratory for Information \& Decision Systems~(LIDS), Massachusetts Institute of Technology, Cambridge, MA 02139, USA. Email: {\tt\scriptsize \{shapelim, jn.shi, lcarlone\}@mit.edu}. \hfill \break
  \indent $^{2}$The School of Electrical Engineering, KAIST (Korea Advanced Institute of Science and Technology), Daejeon, 34141, Republic of Korea. Email: {\tt\scriptsize \{ted97k, gunhee\_shin, hmyung\}@kaist.ac.kr}. \hfill \break
  \indent $^{3}$Dexory, UK. E-mail: {\tt\scriptsize ignaciovizzo@gmail.com}. \hfill \break
  \indent $^{4}$Computer Science Engineering and Interdisciplinary Program of AI, Seoul National University, Seoul, 08826, Republic of Korea. Email:~{\tt\scriptsize jaesik.park@snu.ac.kr}.  \hfill \break
  \indent This work was partially supported by the Basic Science Research Program through the National Research Foundation of Korea (NRF) funded by the Ministry of Education (RS-2024-00415018, 30\%) and the NRF grant funded by the Korea government (MSIT)\,(No.~RS-2024-00461409, 30\%). The Korean students were supported by the BK21 FOUR in Republic of Korea (10\%). Jaesik Park was supported by MSIT grant (RS-2021-II211343: AI Graduate School Program at Seoul National University (5\%) and 2023R1A1C200781211 (25\%).}
}

\begin{document}
\maketitle
\thispagestyle{empty}
\pagestyle{empty}

\begin{abstract}
	While global point cloud registration systems have advanced significantly in all aspects, many studies have focused on specific components, such as feature extraction, graph-theoretic pruning, or pose solvers. 
	In this paper, we take a holistic view on the registration problem and develop an open-source and versatile C++ library for point cloud registration, called \textit{KISS-Matcher}.  
	\textit{KISS-Matcher} combines a novel feature detector, \textit{Faster-PFH}, that improves over the classical fast point feature histogram (FPFH). Moreover, it adopts a $k$-core-based graph-theoretic pruning to reduce the time complexity of rejecting outlier correspondences.
	Finally, it combines these modules in a complete, user-friendly, and ready-to-use pipeline.
	As verified by extensive experiments, KISS-Matcher has superior scalability and broad applicability, achieving a substantial speed-up compared to state-of-the-art outlier-robust registration pipelines while preserving accuracy. Our code will be available at \href{https://github.com/MIT-SPARK/KISS-Matcher}{\texttt{https://github.com/MIT-SPARK/KISS-Matcher}}.
\end{abstract}

\section{Introduction}
\label{sec:intro}


3D point cloud registration, which estimates the relative pose between two partially overlapping point clouds, is a fundamental problem in robotic and computer vision, and arises in the context of localization, mapping, and object pose estimation~\cite{Yang20tro-teaser, Lim24ijrr-Quatropp, Lim22icra-Quatro, Zhou16eccv-FastGlobalRegistration, Yang16pami-goicp, Bernreiter21ral-PHASER,Yin23icra-Segregator}.
Unlike iterative closest point~(ICP)-based approaches~\cite{Segal09rss-GeneralizedICP, Oomerleau12ijrr-ethpc, Pomerleau13auro-ICPcomparison, Pomerleau14icra-LongTerm3DMap, Koide21icra-VGICP,Vizzo23ral-KISSICP}, which have narrow convergence regions and only converge to good estimates when a suitable initial guess is available,
\emph{global} registration aims at estimating the relative pose without requiring an initial guess.
For this reason, global registration is widely used in applications~\cite{Kim18icra-RobustLoc, Aoki24icra-3DBBS,Yin24ijcv-LiDARLocSurvey,Chen22ar-Overlapnet,Cattaneo22tro-LCDNet,Lee22arxiv-LearningReg}, where it typically provides an initial alignment
to enable ICP-based methods to converge to better solutions.

\begin{figure}[t!]
	\centering
	\captionsetup{font=footnotesize}
	\begin{subfigure}[b]{0.45\textwidth}
	\includegraphics[width=1.0\textwidth]{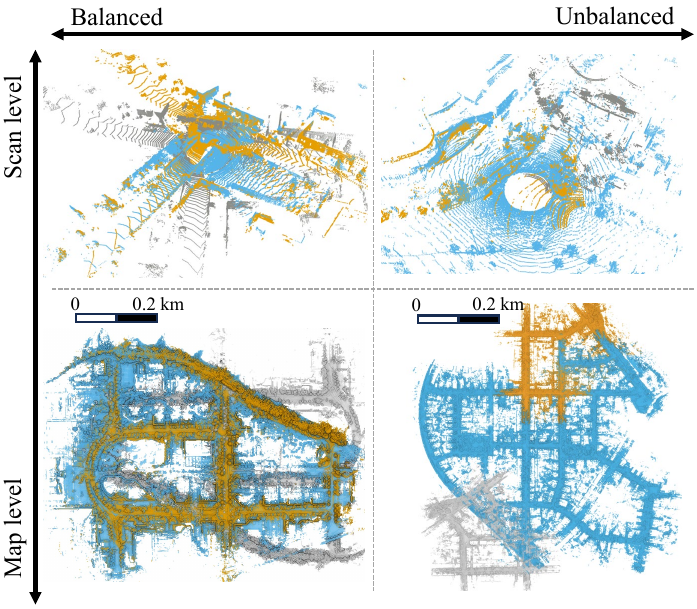}
	\caption{}
	\end{subfigure}
	\begin{subfigure}[b]{0.45\textwidth}
	\includegraphics[width=1.0\textwidth]{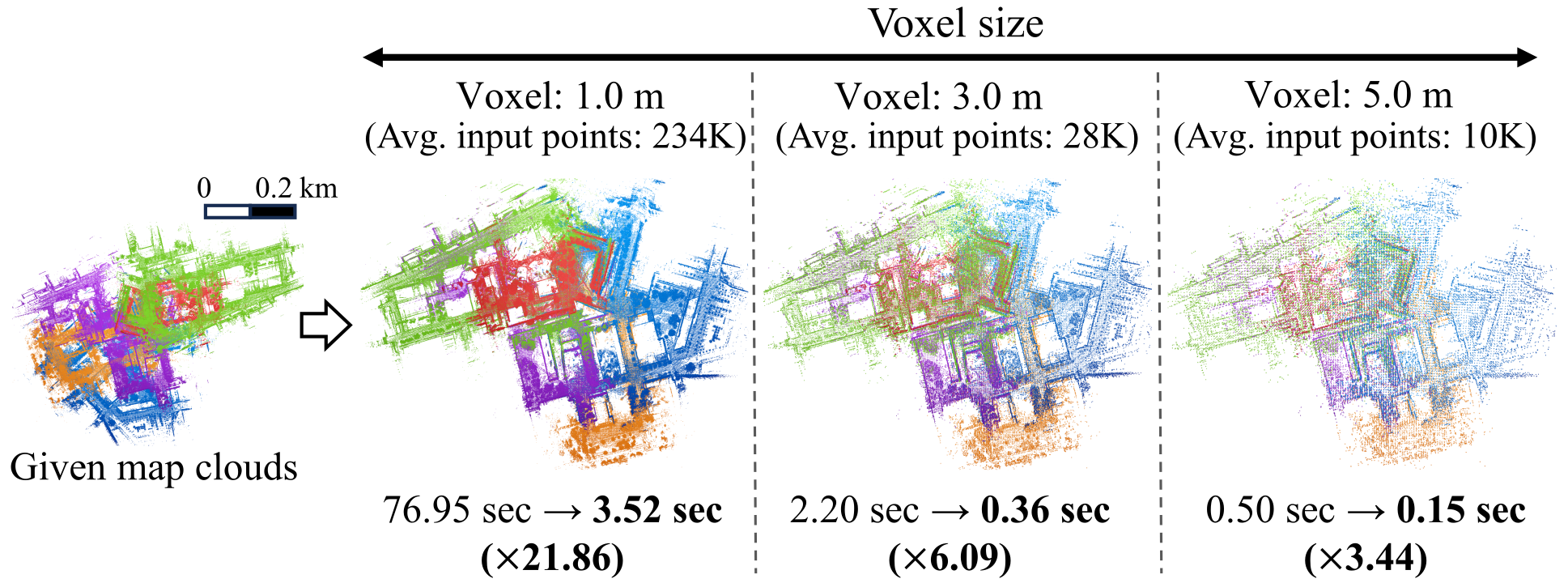}
	\caption{}
	\end{subfigure}
	\caption{(a)~Registration results of the proposed approach, called \textit{KISS-Matcher}. KISS-Matcher can be applied to both balanced and unbalanced settings, as well as egocentric scan-level and map-level scales.
	Gray and cyan colors indicate source and target clouds, respectively, and yellow color indicates the warped source cloud by our approach.
	(b)~Speed comparison between the TEASER++ pipeline~\cite{Yang20tro-teaser} and our KISS-Matcher. Note that the time includes the entire pipeline, from feature extraction and matching to graph-theoretic pruning and pose estimation.
	Note that our approach can achieve a substantial speed-up with respect to TEASER++ and its superiority becomes more pronounced as the number of given points increases. Different colors represent different sessions~(best viewed in color).}
	\label{fig:main_fig}
  \vspace{-0.4cm}
\end{figure}

Despite remarkable progress over the past decade, prior works have often focused on enhancing performance of specific components (\eg feature extraction, graph- theoretic pruning, or pose solvers), often disregarding real-time capabilities and broader applicability of the overall registration pipeline.
In particular, while recent deep learning-based approaches address generalization in untrained scenes~\cite{Ao21cvpr-Spinnet,Poiesi22pami-GeDi,Ao23CVPR-BUFFER},
some learning-based methods overlook the time required for data preprocessing~\cite{Ao23CVPR-BUFFER} or their substantial reliance on millions of random sample consensus~(RANSAC) iterations~\cite{Yew18eccv-3dfeatnet,Choi19iccv-FCGF,Huang21cvpr-PREDATORRegistration,Poiesi22pami-GeDi}, which can take a few to tens of seconds.
Furthermore, not only deep learning methods but also some traditional approaches are often limited to scan-level registration and do not scale to submap-level or map-level registration, which restricts their applicability~\cite{Shan21icra-RobustPPR,Guadagnino23ral-FastSparseLO,Chen21icra-RangeImageLiDARLoc}; therefore, it is desirable to develop \emph{scalable} registration methods that are applicable to both the scan level and the large-scale map level, as shown in \figref{fig:main_fig}(a).

In this paper, we propose a fast and scalable registration system, that combines advances in each component of the registration pipeline into a versatile and ready-to-use C++ library.
Sharing the philosophy of KISS-ICP~\cite{Vizzo23ral-KISSICP}, which focuses on applicability and attempts at reducing the reliance on
 sensor-specific assumptions or laborious parameter tuning,
we propose a ``keep it simple and scalable'' registration pipeline called \textit{KISS-Matcher}.
To this end, we revisit our research on outlier-robust registration~\cite{Yang20tro-teaser, Lim24ijrr-Quatropp, Lim22icra-Quatro, Yang20ral-GNC,Tzoumas19iros-outliers} and focus on making the entire registration pipeline more efficient and ready to use, without sacrificing robustness.

We approach the registration problem from a holistic perspective, specifically addressing the issues in the TEASER++ pipeline~\cite{Yang20tro-teaser}, and in particular its lack of scalability to a large number of correspondences, which is common in map-level matching scenarios.
To this end, we boost the speed of a classical handcraft descriptor, fast point feature histogram~(FPFH)~\cite{Rusu09icra-fast3Dkeypoints}, by proposing \textit{Faster-PFH}, and adopt a $k$-core-based graph-theoretic pruning to reject spurious correspondences.
Our system is several times faster than TEASER++ while achieving similar performance (\figref{fig:main_fig}(b)).

Extensive experiments demonstrate that KISS-Matcher (i)~is on par with state-of-the-art approaches, including learning-based methods, in scan-level registration,
(ii)~particularly exhibits superior applicability and scalability in submap-level and map-level registration,
and (iii)~successfully resolves the computational bottlenecks in~\cite{Yang20tro-teaser}.

\section{Related Work}
\label{sec:related}

\begin{figure}[t!]
 	\centering
 	\begin{subfigure}[b]{0.41\textwidth}
 		\includegraphics[width=1.0\textwidth]{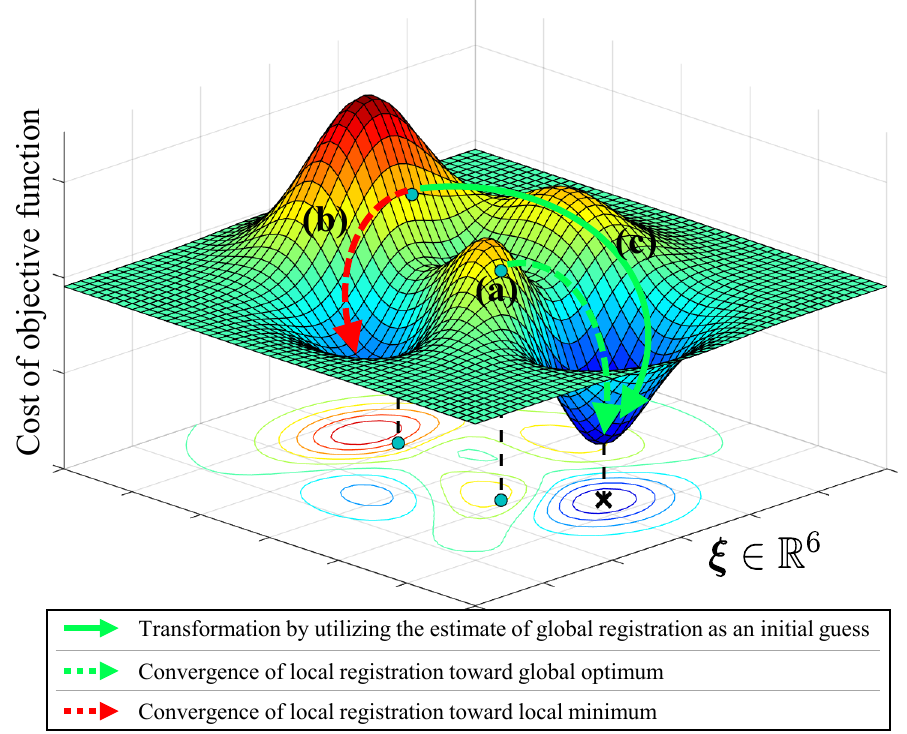}
 	\end{subfigure}
 	\captionsetup{font=footnotesize}
 	\caption{Visual description of the convergence direction of local and global registration, where the space is parameterized by 3D twist, $\zeta \in \mathbb{R}^6$. (a) Once the pose discrepancy between the source and target is not large, the local registration approaches are highly likely to converge to a global optimum. (b)~However, as the pose discrepancy increases, the local registration approaches are likely to converge to a local minimum (red dashed arrow), (c) so global registration is necessary to help local registration converge to the global optimum by initially reducing the viewpoint discrepancy~(green solid arrow).}
 	\label{fig:local_global_registration_optimum}
  \vspace{-0.2cm}
\end{figure}

Point cloud registration methods are commonly classified into two categories depending on their convergence capabilities: local registration~\cite{Besl92pami, Segal09rss-GeneralizedICP,Pomerleau13auro-ICPcomparison,Koide21icra-VGICP} and global registration methods~\cite{Fischler81,Dong17isprsremotesensing-GHICP,Yang16pami-goicp,Zhou16eccv-FastGlobalRegistration,Bernreiter21ral-PHASER}.
When two point clouds are given, local registration methods, \eg based on ICP and variants~\cite{Vizzo23ral-KISSICP}, can converge to the global optimum when the pose discrepancy between the point cloud is small (or, equivalently, when an initial guess for the relative pose is available); see \figref{fig:local_global_registration_optimum}(a).
However, in the presence of a large pose discrepancy, local registration can fail to converge to the optimum~(red dashed arrow in \figref{fig:local_global_registration_optimum}(b)) as the viewpoint difference becomes larger.
Unlike local registration, global registration enables robust registration despite large pose discrepancies~(green solid arrow in \figref{fig:local_global_registration_optimum}(c)).
Therefore, our global registration pipeline aims at computing a fast and robust initial alignment, which can be possibly used to bootstrap local registration methods.

Global registration methods can be further categorized into two approaches: a)~correspondence-based~\cite{Fischler81,Yang16pami-goicp,Zhou16eccv-FastGlobalRegistration,Dong17isprsremotesensing-GHICP,Lei17tip-FastDescriptors} and b)~correspondence-free approaches~\cite{Rouhani11iccv-CorrespondenceFreeReg,Brown19pr-AFamiliyofBnB,Bernreiter21ral-PHASER, Papazov12ijrr-Rigid3DGeometryMatching,Chum03jprs-LocallyOptimizedRANSAC,Choi97jcv-RANSAC,Schnabel07cgf-EfficientRANSAC,Olsson09pami-bnbRegistration, Hartley09ijcv-globalRotationRegistration, Pan19robotbiomim-MultiViewBnB}.
Our study primarily focuses on the former category because correspondence-based approaches are typically faster than correspondence-free methods and offer better applicability if the correspondences are well-estimated~\cite{Zhou16eccv-FastGlobalRegistration}.

In recent years, numerous researchers have demonstrated that outlier-robust registration approaches can endure up to~99\% of outliers, while showing sufficiently fast speed for scan-level registration~\cite{Yang20tro-teaser,Yang20ral-GNC,Lim24ijrr-Quatropp,Sun21ral-RANSIC,Lee24arxiv-PCR99,Sun22ral-TriVoC}.
However, these approaches have primarily focused on the solver aspect of registration and have only been tested on object-level or scan-level registration problems. 
Indeed, when testing these methods, \eg~\cite{Yang20tro-teaser},  on submap-level or map-level registration problems,
 we observe a substantial slowdown that makes these approaches less appealing (\figref{fig:main_fig}(b)).


Deep learning-based approaches have recently shown promising performance~\cite{Zeng17cvpr-3dmatch,Wang19ICCV-DeepClosestPoint, Gojcic20CVPR-learnMultiviewRegistration, Choy20cvpr-deepGlobalRegistration,Poiesi21icpr-DIP,Huang21cvpr-PREDATORRegistration,Ao21cvpr-Spinnet,Bai20cvpr-D3Feat,Poiesi22pami-GeDi} by significantly improving the expressiveness of feature descriptors and reducing the outlier ratio within correspondences.
However, they tend to overfit to their training datasets and occasionally fail when applied to data captured by different sensor configurations or in untrained scenarios~(see \secref{sec:scalability}).

The key observation behind this work is that previous outlier-robust registration approaches, \eg~\cite{Yang20tro-teaser}, have demonstrated 
impressive performance in scan-level registration even when applied to handcrafted feature matching techniques, such as FPFH~\cite{Rusu09icra-fast3Dkeypoints}. At the same time, when integrated in a complete registration system, these pipelines have two main bottlenecks. 
First, the feature computation remains expensive for large-scale submap-level or map-level registration problems.
Second, the graph-theoretic approach used in~\cite{Yang20tro-teaser} becomes relatively slow for registration problems with more than a few thousand correspondences. 
Therefore, this study returns to the classical handcraft descriptor, FPFH~\cite{Rusu09icra-fast3Dkeypoints}, and speeds up FPFH by minimizing unnecessary computations as much as possible.
Furthermore, we revisit the graph-theoretic matching process and use $k$-core-based graph theoretic pruning to reduce time complexity, based on~\cite{Shi21icra-robin}.




\section{KISS-Matcher: Robust, Fast, and Scalable Outlier-Robust Registration}
\label{sec:main}

In this section, the pipeline of our KISS-Matcher is presented, following the typical pipeline of feature-based outlier-robust registration as illustrated in \figref{fig:overview}.
Our KISS-Matcher is composed of four components: a)~geometric suppression, b)~Faster-PFH-based feature extraction and initial matching,
c)~$k$-core-based graph-theoretic outlier rejection, and d)~graduated-non-convexity-based non-minimal solver.

\subsection{Problem Definition}\label{subsec:problem}
\newcommand{\corr}{\mathcal{A}}
\newcommand{\estoutliers}{\hat{\mathcal{O}}}
\newcommand{\srcpt}{\srcpoint_\srcidx}
\newcommand{\tgtpt}{\tgtpoint_\tgtidx}

Our objective is to align two unordered voxelized point clouds with a voxel size $v$, namely the source~$\srccloud$ and target~$\tgtcloud$ point clouds.
To this end, we establish correspondences between the two point clouds, which is followed by robust estimation to suppress the undesirable effect of outliers. 

Formally, let us assume that the $\kth$-th pair (or the $\kth$-th correspondence) obtained through matching consists of the 3D point $\srcpoint_\srcidx \in \srccloud$ and the 3D point $\tgtpoint_\tgtidx \in \tgtcloud$.
Then, the $k$-th measurement can be modeled as follows:

\begin{equation}
 \label{eq:p-reg-gen-model}
 \tgtpoint_\tgtidx = \MR \srcpoint_\srcidx + \vt + \vepsilon_{\srcidx, \tgtidx},
\end{equation}

\noindent where $\MR \in \SOthree$ is the relative rotation matrix, $\vt\in\mathbb{R}^{3}$ is translation vector, and $\vepsilon_{\srcidx,\tgtidx} \in \Real{3}$ is the measurement noise.
Finally, by denoting the initial correspondence set as $\corresp$, our objective function can be expressed as:

\begin{figure}[t!]
 \centering
 \begin{subfigure}[b]{0.48\textwidth}
  \includegraphics[width=1.0\textwidth]{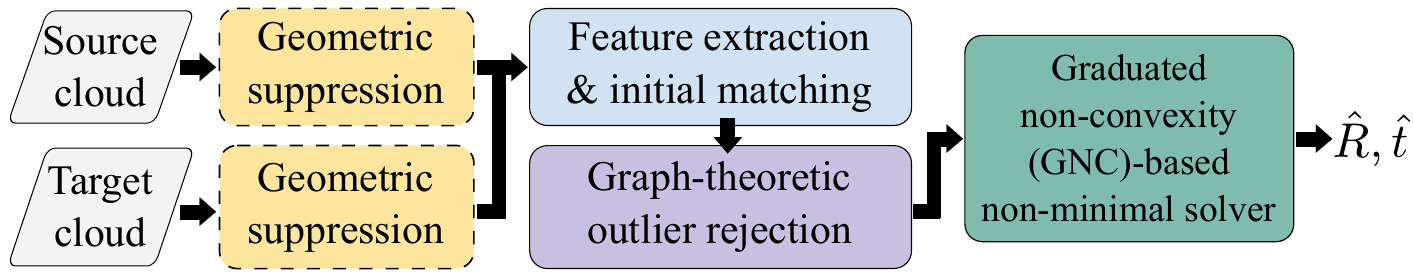}
 \end{subfigure}
 \captionsetup{font=footnotesize}
 \caption{Pipeline of the generic outlier-robust registration framework~\cite{Yang20tro-teaser,Lim24ijrr-Quatropp}.
  Especially, our KISS-Matcher comprises four components: (i) geometric suppression for both source and target clouds,
  (ii)~faster point feature histogram~(Faster-PFH)-based feature extraction and initial matching,
  (iii)~$k$-core-based graph-theoretic outlier rejection,
  and (iv)~graduated non-convexity~(GNC)-based non-minimal solver.}
 \label{fig:overview}
\end{figure}

\begin{equation}
\hat{\MR}, \hat{\vt} =\argmin_{\MR \in \mathrm{SO}(3), \vt \in \mathbb{R}^{3}}  \sum_{(i,j) \in  \corr \setminus \estoutliers} \rho\Big( \twonorm{\tgtpoint_\tgtidx -\MR \srcpoint_\srcidx -\vt} \Big),
 \label{eqn:final_goal}
\end{equation}

\noindent where $\rho(\cdot)$ denotes a robust loss function intended to suppress undesirable large errors caused by false correspondences~(or outliers),
and $\hat{\mathcal{O}}$ denotes gross outliers which have been pre-filtered before the optimization (\eg by the graph-theoretic outlier pruning in \figref{fig:overview}).
Thus, \eqref{eqn:final_goal} attempts to estimate the relative pose between the source and target point cloud while being robust to outliers. 

\subsection{Geometric Suppression as a Preprocessing Step}\label{subsec:kissmatcher}

As reported by Yang\etalcite{Yang20tro-teaser}, correspondences resulting from repeating patterns in the environment, such as points on the ground, ceiling, or planar walls of a building, create an overwhelming amount of outliers (\eg when many pairs of different points on a plane are incorrectly identified as matching), which hinders the quality of the pose estimates from~\eqref{eqn:final_goal}. 
Thus, if these points can be easily detected, they should be filtered out in a preprocessing step to make our approach more robust~\cite{Lim24ijrr-Quatropp}; we name this preprocessing \textit{geometric suppression}.
Consequently, this geometric suppression not only prevents the subsequent steps from converging to poor solutions but also helps  enhance the distinctiveness of features by rejecting redundant and non-distinctive points in advance.
In addition, it can significantly speed up the subsequent steps by reducing the number of cloud points~(see \secref{sec:runtime}).
To maintain generality, we continue to denote with $\srccloud$ and $\tgtcloud$ the filtered point clouds after the geometric suppression has been applied.

\subsection{Faster-PFH: Boosting FPFH Speed}\label{subsec:fpfh}
\newcommand{\querypoint}{\srcpoint_q}
\newcommand{\pfhi}{a}
\newcommand{\pfhj}{b}
\newcommand{\histidx}{l}
\newcommand{\rnormal}{r_\text{normal}}
\newcommand{\rfpfh}{r_\text{FPFH}}
\newcommand{\Pfpfh}{\srccloud_\text{FPFH}}
\newcommand{\Pnormal}{\srccloud_\text{normal}}
\newcommand{\Pval}{\srccloud_\text{val}}
\newcommand{\Pvalid}{\srccloud_\text{valid}}

\begin{figure}[t!]
 \captionsetup{font=footnotesize}
 \centering
 \begin{subfigure}[b]{0.48\textwidth}
  \includegraphics[trim=5 0 0 0,clip,width=1.0\textwidth]{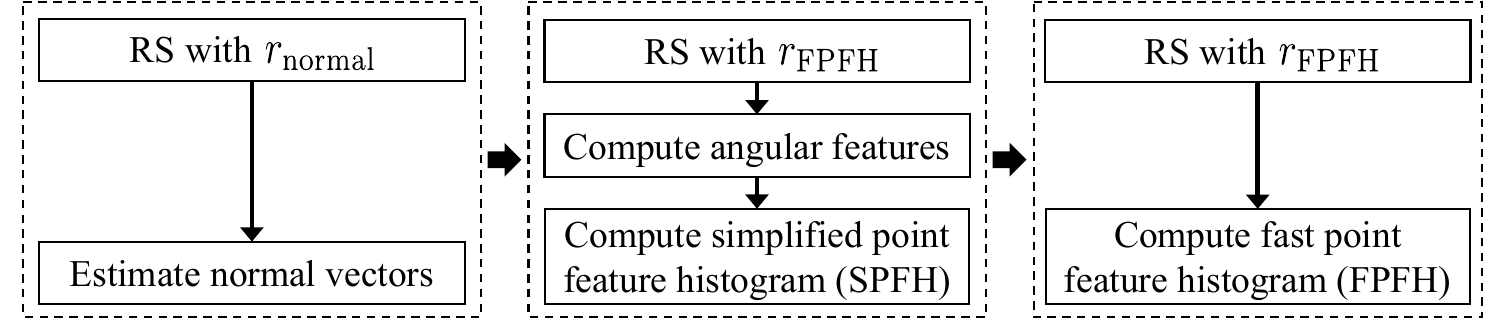}
  \caption{}
  \vspace{0.1cm}
 \end{subfigure}
 \begin{subfigure}[b]{0.48\textwidth}
  \includegraphics[trim=5 0 0 0,clip,width=1.0\textwidth]{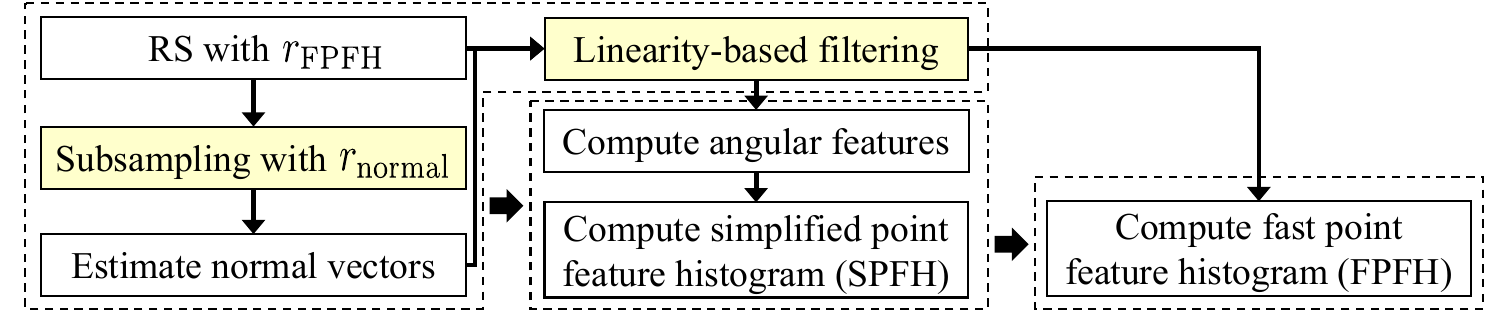}
  \caption{}
 \end{subfigure}
 \caption{Block diagram of (a)~fast point feature histograms~(FPFH) and (b)~our proposed Faster-PFH.
 RS is an abbreviation of radius search. Unlike FPFH, our Faster-PFH performs the radius search only once for both normal estimation and feature extraction, and filters out noisy points to reduce computational cost as much as possible.}
 \label{fig:fpfh_and_ours}
\end{figure}

Next, we need to extract and match features between the point clouds $\srccloud$ and $\tgtcloud$.
Towards this goal, we propose \textit{Faster-PFH}, which is a more efficient variant of FPFH~\cite{Rusu09icra-fast3Dkeypoints} while retaining similar performance.
As illustrated in \figref{fig:fpfh_and_ours}(a), the process of FPFH computation mainly follows three steps:
(i)~normal estimation for each point using neighboring points within the radius $\rnormal$,
(ii)~angular feature extraction between a query point and neighboring points with the FPFH radius $\rfpfh$, which is followed by computing the simplified PFH~(SPFH) signatures based on the distribution of the angular features, 
and (iii)~computing FPFH signatures by weighted averaging of SPFH of neighboring points~\cite{Rusu09icra-fast3Dkeypoints}.

%

However, existing feature extraction of FPFH may not be suitable in real-time applications; for instance, applying FPFH to the voxelized source and target clouds captured by a 64-channel LiDAR sensor, whose numbers of points range from 10K to 30K, takes more than 0.1 seconds, even with multi-threading on an Intel(R) Core(TM) i9-13900 CPU; see \secref{sec:runtime}.
In particular, FPFH has two computational bottlenecks. 
First, the computational inefficiency arises from multiple executions of the neighboring point search~(\ie three rounds of radius search, referred to as ``RS with $r_\text{normal}$'' or ``RS with $r_\text{FPFH}$'' in \figref{fig:fpfh_and_ours}(a)).
Second, because no thorough check exists to determine whether their normal estimation is reliable or not owing to its decoupled scheme,
SPFH and FPFH extraction are unnecessarily performed on points with unreliable normal vectors, leading to a degradation in the expressiveness of neighboring FPFH features.

To address these two issues, we adopt two strategies.
First, to reduce unnecessary computation cost,
we perform the neighboring point search only once for each point with $\rfpfh$ and reuse the results.
Then, based on the fact that $\rnormal \leq \rfpfh$, we subsample neighboring points for normal estimation, $\srccloud_{\text{normal}}$, from the outputs of the neighboring search with $\rfpfh$, $\srccloud_{\text{FPFH}}$~(\ie~$\srccloud_{\text{normal}} \subset \srccloud_{\text{FPFH}}$). 

Second, for each point, if either the cardinality of $\Pfpfh$ is smaller than $\tau_{\text{num}}$ or the linearity~(\ie $\frac{\lambda_{1}-\lambda_{2}}{\lambda_{1}}$, where $\lambda_{1}$, $\lambda_{2}$, and $\lambda_{3}$ are three eigenvalues obtained from principal component analysis~(PCA) for normal estimation using $\Pnormal$, satisfying $\lambda_{1} \geq \lambda_{2} \geq \lambda_{3}$) is higher than~$\tau_{\text{lin}}$,
the point is excluded in computation of angular features, SPFH, and FPFH features.
Here, $\tau_{\text{num}}$ and $\tau_{\text{lin}}$ are user-defined thresholds.
By doing so, we only generate descriptors for the reliable points, similar to semi-direct methods~\cite{Forster14icra}.

Then, the initial correspondence set $\mathcal{A}$ is established through mutual matching (or reciprocity test~\cite{Zhou16eccv-FastGlobalRegistration}) by using the outputs of Faster-PFH.
Consequently, we can not only generate more reliable descriptors but also reduce computational costs by preemptively rejecting potential points that would otherwise lead to outliers.
As a result, we improved the speed by approximately 4.5~times and 2.4~times in the single-threaded and multi-threaded cases, respectively, while maintaining performance~(see \secref{sec:runtime}).

\subsection{$k$-Core-Based Graph-Theoretic Outlier Pruning}\label{subsec:robin}

Next, we resolve the empirical problem where the existing TEASER++ pipeline becomes drastically slower as the number of correspondences increases, which frequently occurs in submap-level or map-level registrations and therefore hinders the scalability of the pipeline.
This slowdown is due to the maximum clique inlier selection (MCIS) in TEASER++, which has exponential time complexity 
and empirically becomes slow once the number of correspondences exceeds 500-1,000 points.
Here, we use $k$-core-based graph-theoretic pruning~\cite{Shi21icra-robin} to circumvent the exponential complexity.

To be more concrete, our pruning process mainly follows two steps to reject as many spurious correspondences as possible.
First, by taking $\mathcal{A}$ as an input, we exploit the concept of \textit{pairwise-invariants}~\cite{Zhou16eccv-FastGlobalRegistration,Shi21icra-robin}, which is a specialized version of $n$-invariants for point cloud registration~\cite{Shi21icra-robin}.
Formally, let us consider two inlier correspondence in $\corr$, with indices as $(i, j)$ and $(i^\prime, j^\prime)$, respectively.
 Assuming the inlier noise is bounded (\eg $\|\vepsilon_{i,j}\|\leq\!\beta$ and $\|\vepsilon_{i^\prime, j^\prime}\|\leq\!\beta$, where $\beta>0$ is a user-defined threshold), one can use \eqref{eq:p-reg-gen-model} to establish that the inlier correspondences have to satisfy the following inequality (see~\cite{Yang20tro-teaser} for a derivation):

\beq
\label{eq:test-p-ref3}
-2\beta  \leq  \| \vb_\tgtidx - \vb_{\tgtidx^\prime} \| - \| \va_\srcidx  - \va_{\srcidx^\prime} \| \leq  2\beta.
\eeq
Based on \eqref{eq:test-p-ref3}, we represent the relationships of the measurements as a compatibility graph $\mathcal{G}(\mathcal{V}, \mathcal{E})$,
where each vertex in the vertex set $\mathcal{V}$ represents a correspondence pair in $\mathcal{A}$ and each edge in the edge set $\mathcal{E}$ is established between two vertices if their corresponding measurements satisfy the consistency relation \eqref{eq:test-p-ref3}.

Second, by finding large sets of mutually consistent measurements, we can identify potential inliers and filter out gross outliers.
In particular, once $\mathcal{G}(\mathcal{V}, \mathcal{E})$ is constructed, we look for the maximum $k$-core of the graph 
to obtain a large set of compatible measurements.
Consequently, non-compatible correspondences are discarded as outliers (this is the set~$\estoutliers$ in \eqref{eqn:final_goal}).
As shown in Shi\etalcite{Shi21icra-robin}, finding the maximum $k$-core approximates the performance of MCIS but operates with a linear time complexity of $O(|\mathcal{V}|+|\mathcal{E}|)$.
In contrast, the fastest algorithm for identifying the maximum clique in arbitrary graphs is $O(1.1888^{|\mathcal{V}|})$~\cite{Robson86jalg-MaximumClique, Robson01techreport-FindingMaximumClique,Rossi15parallel}, whose complexity grows exponentially as $|\mathcal{V}|$ increases.
In addition to leveraging $k$-core-based pruning, we employ the compressed sparse row~(CSR) format instead of the adjacency matrix format used in the original TEASER++ pipeline, because the CSR format is a more memory-efficient for large sparse graphs.

Nevertheless, if too many correspondences are given, a slowdown is unavoidable as our method also experiences a linear increase in runtime.
To address this issue, before performing the above two steps, we first select the top $N_\tau$ correspondences with the lowest descriptor distance ratios, motivated by the ratio test~\cite{lowe2004ijcv-distinctive}.
By doing so, this ratio-based filtering rejects pairs with low feature distinctiveness in advance, ensuring that the number of correspondences does not exceed $N_\tau$.

\subsection{Graduated Non-Convexity-Based Non-Minimal Solver}\label{subsec:nonminimal_solver}

Note that while the graph-theoretic outlier pruning typically filters out gross outliers, some outliers might remain in the set $\corr \setminus \estoutliers$. Therefore, even after $\estoutliers$ is rejected, we use the graduated non-convexity (GNC)~\cite{Yang20ral-GNC} solver to robustly estimate the relative pose by solving \eqref{eqn:final_goal}.

A useful feature of the GNC-based solver is that it allows us to determine whether the registration is valid by checking if the cardinality of the final inlier set, $\mathcal{I}_\text{final}$, is sufficiently large~\cite{Yang20tro-teaser}.
By doing so, contrary to other solvers, \eg based on learning-based approaches, our GNC-based solver enables to filter out failed and potentially inaccurate registration results, thereby increasing registration precision~(\figref{fig:mulran_comparison}(a)).

\section{Experimental Evaluation}
\label{sec:exp}

We perform an extensive evaluation of KISS-Matcher across various robotics scenarios, ranging from scan-level to map-level registration.
Our results show that  KISS-Matcher (i)~is on par with state-of-the-art approaches in the scan-level registration
yet (ii) has superior scalability and applicability not only in submap-level but also map-level registration, and
(iii)~our Faster-PFH and $k$-core-based matching in KISS-Matcher are faster than the existing FPFH and graph-theoretic pruning in the TEASER++ library~\cite{Yang20tro-teaser}.

\subsection{Experimental Setup}
\newcommand{\btwdist}{\sim}

In the experiments, we use the KITTI~\cite{Geiger13ijrr-KITTI} and the MulRan~\cite{kim20icra-MulRan} datasets to evaluate the scan-level registration in cases where the viewpoints between the source and target clouds are distant.
In particular, because global registration is vital for loop closing in LiDAR-based SLAM, we use the loop closing benchmark~\cite{Lim24ijrr-Quatropp}.
Thus, the \textit{$A \btwdist B$\; loop closing test} in this paper refers to the case where we sample source and target pairs with a sample size of 1,000, and the distance between the ground truth source and target poses lies between $A$ and $B$. In addition, these source and target pairs have a sufficiently large frame difference~(more details are explained in Lim\etalcite{Lim24ijrr-Quatropp}).

Furthermore, to test the scalability, we use submap-level and map-level point clouds.
For the submap clouds, $W$ indicates the submap window size. For instance,~$W=3$ means that $\{n-1, n, n+1\}$ frames with their poses are used to build the submap cloud, where $n$ is the arbitrary $n$-th frame.
For map-level clouds, we used the outputs of FAST-LIO2-based SLAM~\cite{xu22tro-FastLIO2} in the Kimera-Multi dataset~\cite{Tian23iros-KimeraMultiExperiments}; each map-level cloud corresponds to a different robot trajectory in the multi-robot dataset.
Note that those map clouds are hundreds of meters in scale and are hundred times larger than a LiDAR scan in the KITTI benchmark; moreover, the corresponding point clouds have inherent pose errors and noise, hence further stress-testing our approach.


\begin{figure}[t!]
	\captionsetup{font=footnotesize}
	\centering
	\begin{subfigure}[b]{0.23\textwidth}
		\includegraphics[width=1.0\textwidth]{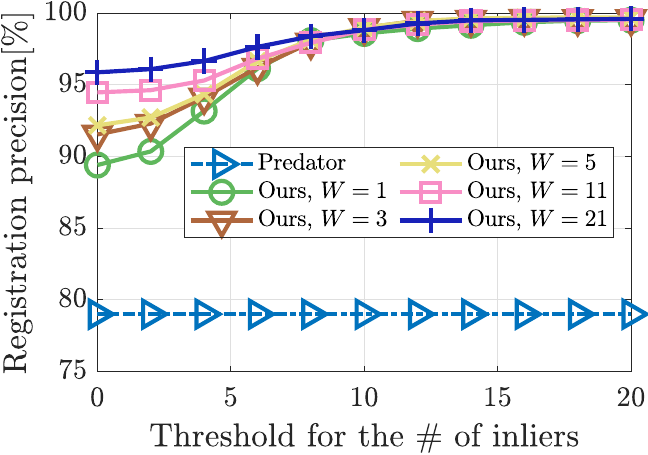}
		\caption{}
	\end{subfigure}
	\begin{subfigure}[b]{0.23\textwidth}
		\includegraphics[width=1.0\textwidth]{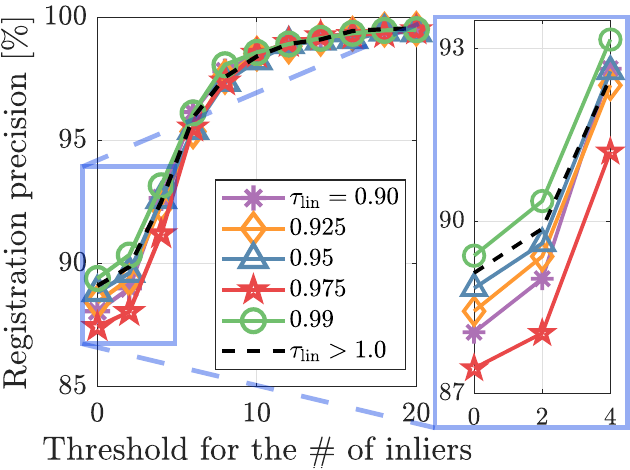}
		\caption{}
	\end{subfigure}
  \caption{Average registration precision with respect to the threshold on the number of inliers in the $2 \sim 12$\;m loop closing test~\cite{Lim24ijrr-Quatropp} in the \texttt{Riverside} sequences of the MulRan dataset~\cite{kim20icra-MulRan}.
	Note that, by checking the number of final inliers $\mathcal{I}_\text{final}$, our approach can easily reject failure cases.
		(a)~Performance comparison with Predator~\cite{Huang21cvpr-PREDATORRegistration} trained in the KITTI dataset to compare the generalization capabilities. Note that performance degradation of Predator occurred when tested on the MulRan dataset, despite both datasets being captured with a 64-channel LiDAR sensor but having different laser ray patterns. Registration precision is defined as $\frac{\text{\# of successful registrations}}{\text{\# of pairs considered as valid}}$.
		(b) Ablation studies with different thresholds of linearity,~$\tau_\text{lin}$.}
	\label{fig:mulran_comparison}
	\vsfig
\end{figure}

As a quantitative metric, we used the success rate, which is a crucial metric to directly evaluate the robustness of global registration~\cite{Lim24ijrr-Quatropp}.
Thus, the registration is considered successful if both translation and rotation errors are below 2\;m and 5$^\circ$, respectively~\cite{Yew18eccv-3dfeatnet}.
For successful registration results, the relative translation error (RTE) and relative rotation error~(RRE) were used, which are defined as follows:

\newcommand{\numsuc}{N_\text{success}}
\begin{itemize}
	\item $\text{RTE}= \sum_{n=1}^{\numsuc} (\vt_{n, \text{GT}}-{\hat{\vt}}_{n})^{2} / \numsuc$,
	\item $\text{RRE}= \frac{180}{\pi} \sum_{n=1}^{\numsuc} | \cos^{-1} (\frac{\operatorname{Tr}\left({\hat{\MR}}_{n}^{\intercal} \MR_{n, \text{GT}}\right)-1}{2}) | / \numsuc $
\end{itemize}

\noindent where $\vt_{n, \text{GT}}$ and $\MR_{n, \text{GT}}$ are the $n$-th ground truth translation and rotation, respectively;
$\numsuc$ denotes the number of successful registration results.

\subsection{Parameters of KISS-Matcher}

One of the characteristics of our KISS-Matcher is that all variables can be parameterized in terms of the voxel size $v$ for better usability.
In particular, we set $r_\text{normal}=3.5v$, $r_\text{FPFH}=5.0v$, $\tau_{\text{num}}=3$, and $\tau_{\text{lin}}=0.99$ for Faster-PFH,
and $\beta=1.5v$ and $N_\tau = 3,000$ for $k$-core based graph-theoretic pruning.
An analysis of impact of these parameters is presented in \secref{sec:runtime}.
\newcommand{\graysetup}{\color{gray!70}}
\begin{table}[t!]
    \captionsetup{font=footnotesize}
    \centering
    \caption{The relative translation error~(RTE), relative rotation error~(RRE), and success rate in the KITTI 10\;m benchmark~\cite{Yew18eccv-3dfeatnet}. 
    $W$ denotes the window size (\ie $W=3$ means that the accumulated cloud with $\{n-1, n, n+1\}$ frames and their poses are used as an input for the $n$-th frame).
    Note that, on an Intel(R) Core(TM) i9-13900 CPU, our KISS-Matcher can operate around 14\;Hz for the entire pipeline (\ie~from feature extraction and matching to pose estimation), whereas other outlier-robust registration pipelines operate at around 6\;Hz.
    In contrast, Predator~\cite{Huang21cvpr-PREDATORRegistration} operates around 0.1\;Hz~(\ie~9.57~sec per pair) for the entire pipeline owing to its heavy reliance on a large number of iterations of RANSAC.}
    \setlength{\tabcolsep}{2pt}
    {\scriptsize
        \begin{tabular}{l|lccc}
            \toprule \midrule
            &{Method} & {RTE [cm]\,$\downarrow$} & {RRE [°]\,$\downarrow$} & {Success rate [\%]\,$\uparrow$} \\
            \midrule
            \parbox[t]{2mm}{\multirow{7}{*}{\rotatebox[origin=c]{90}{Learning-based}}} & 3DFeat-Net~\cite{Yew18eccv-3dfeatnet} & 25.90 &  0.57  & 95.97 \\
            & FCGF~\cite{Choi19iccv-FCGF} & 6.47 &  \hl{0.23} & \hl{99.82} \\
            & DIP~\cite{Poiesi21icpr-DIP}  & 8.69 & 0.44  & 97.30 \\
            & Predator~\cite{Huang21cvpr-PREDATORRegistration} & \hl{5.60} & 0.24  & \hl{99.82} \\
            & SpinNet~\cite{Ao21cvpr-Spinnet} & 9.88  & 0.47  & 99.10 \\
            & D3Feat~\cite{Bai20cvpr-D3Feat} & 11.00  & 0.24  & \hl{99.82} \\
            & GeDi~\cite{Poiesi22pami-GeDi} & 7.55 & 0.33 & \hl{99.82} \\ \midrule
           \parbox[t]{2mm}{\multirow{15}{*}{\rotatebox[origin=c]{90}{Conventional}}}
            & G-ICP~\cite{Segal09rss-GeneralizedICP}  & 8.56  & \hl{0.22} & 37.95 \\
            & STD~\cite{Yuan23icra-STD}, $W=1$  & 26.09 & 0.69 & 19.60 \\
            & STD~\cite{Yuan23icra-STD}, $W=3$  & 20.94 & 0.57 & 30.58 \\
            & STD~\cite{Yuan23icra-STD}, $W=5$  & 23.97 & 0.66 & 33.09 \\
            & MapClosures~\cite{Gupta24icra-MapClosures}, $W=1$  & 38.50  & 1.09  & 69.38  \\
            & MapClosures~\cite{Gupta24icra-MapClosures}, $W=3$  & 31.50  & 0.95  & 82.08  \\
            & MapClosures~\cite{Gupta24icra-MapClosures}, $W=5$  & 32.27  & 1.00  & 86.64  \\
            & FPFH~\cite{Rusu09icra-fast3Dkeypoints} + FGR~\cite{Zhou16eccv-FastGlobalRegistration} & \hl{6.94}  & 0.33  & 98.92  \\
            & FPFH~\cite{Rusu09icra-fast3Dkeypoints} + TEASER++~\cite{Yang20tro-teaser}  & 9.36  & 0.59  & 99.64  \\
            & FPFH~\cite{Rusu09icra-fast3Dkeypoints} + Quatro~\cite{Lim22icra-Quatro}  & 13.15  & 0.94  & 99.64 \\
            & Proposed & 18.10 & 0.94 & \hl{\textbf{100.00}}  \\ \cmidrule(lr){2-5}
            & FPFH + FGR + G-ICP~\cite{Segal09rss-GeneralizedICP} & 1.22     &  0.04  & 99.28  \\ 
            & FPFH + TEASER + G-ICP~\cite{Segal09rss-GeneralizedICP} & \hl{\textbf{1.10}}  &  \hl{\textbf{0.02}}  & 99.64  \\ 
            & FPFH + Quatro + G-ICP~\cite{Segal09rss-GeneralizedICP} & \hl{\textbf{1.10}}  &  0.03  & 99.64  \\ 
            & Proposed + G-ICP~\cite{Segal09rss-GeneralizedICP} & \hl{\textbf{1.10}}  & \hl{\textbf{0.02}} & \hl{\textbf{100.00}}  \\ \midrule \bottomrule
        \end{tabular}
    }
    \label{table:kitti_results}
    \vsfig
\end{table}

\subsection{Evaluation in Scan-Level Registration}\label{sec:performance}

The first experiment evaluates the performance of our approach in scan-to-scan matching cases where the viewpoints between source and target clouds are distant.
In our experiments, we mainly compare our KISS-Matcher with (i)~learning-based approaches~\cite{Yew18eccv-3dfeatnet,Choi19iccv-FCGF,Poiesi21icpr-DIP,Huang21cvpr-PREDATORRegistration,Ao21cvpr-Spinnet,Bai20cvpr-D3Feat,Poiesi22pami-GeDi},
(ii) pose estimation modules in loop closure detection approaches: STD~\cite{Yuan23icra-STD} and MapClosures~\cite{Gupta24icra-MapClosures}, and (iii)~outlier-robust registration approaches~\cite{Zhou16eccv-FastGlobalRegistration,Yang20tro-teaser,Lim22icra-Quatro}.

As presented in \tabref{table:kitti_results},
we demonstrate that our method can robustly provide an initial guess, which is the primary goal of global registration mentioned in \secref{sec:related},
showing that our approach is on par with deep learning-based methods, even without any training procedure.
In particular, our KISS-Matcher also exhibited promising single scan-to-scan registration performance compared with pose estimation modules in STD~\cite{Yuan23icra-STD} and MapClosures~\cite{Gupta24icra-MapClosures}.
This implies that our method has the potential to improve mapping quality when used as a replacement for pose estimation during loop closure in LiDAR-based SLAM systems.

\subsection{Comparison in Terms of Applicability and Scalability}\label{sec:scalability}

While learning-based methods showed promising performance in the training scenes as presented in \tabref{table:kitti_results}, the performance of these approaches is likely to degrade on untrained datasets.
As shown in \figref{fig:mulran_comparison}(a), when comparing with Predator~\cite{Huang21cvpr-PREDATORRegistration}, which was selected because it exhibited the highest RTE and success rate among the learning methods in \tabref{table:kitti_results},
we observed significant performance degradation when tested on the MulRan dataset, despite both datasets being captured with a 64-channel LiDAR sensor but having different laser ray patterns.
In contrast, our method is learning-free, making it less affected by the domain differences of such datasets, showing higher registration precision than Predator.
In addition, we also demonstrate that our approach is easily applicable to submap-level registration, significantly increasing registration precision.
Therefore, this experimental evidence supports our claim that our approach has superior applicability.

In the map-level registration, we observed that KISS-Matcher was the only one to succeed in our experiments, as shown in \figref{fig:map_level_registration}.
Notably, both STD and MapClosures, despite being designed for submap-level registration, failed in map-level registration.
Therefore, we conclude that our method is easily scalable from scan-level to map-level registration and has better scalability compared with other approaches.

\newcommand{\mapscalefigsize}{0.156}
\begin{figure}[t!]
	\captionsetup{font=footnotesize}
	\centering
	\begin{subfigure}[b]{\mapscalefigsize\textwidth}
		\includegraphics[width=1.0\textwidth]{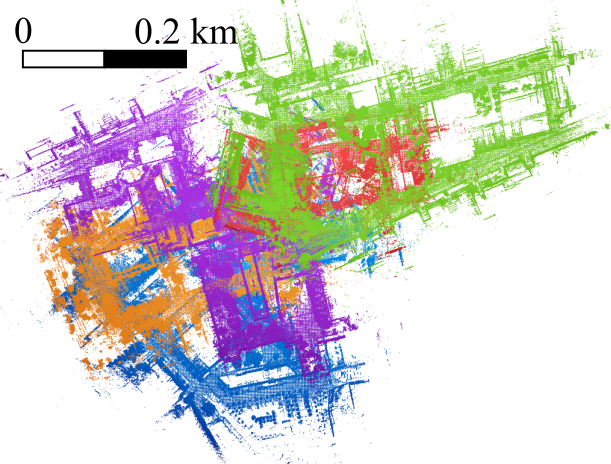}
		\caption{Input}
	\end{subfigure}
	\begin{subfigure}[b]{\mapscalefigsize\textwidth}
		\includegraphics[width=1.0\textwidth]{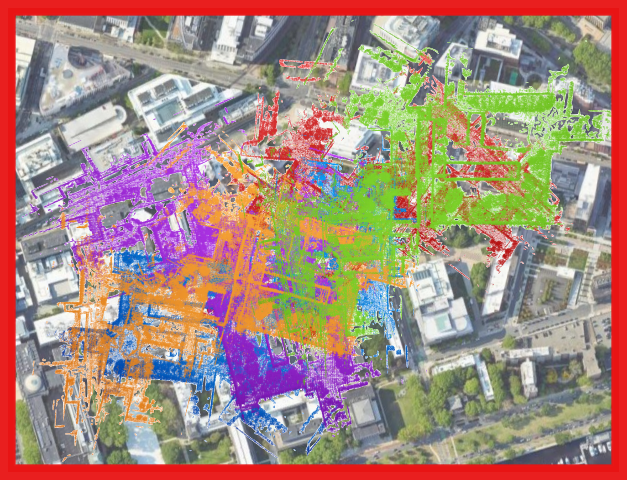}
		\caption{SAC-IA~\cite{Rusu09icra-fast3Dkeypoints}}
	\end{subfigure}
	\begin{subfigure}[b]{\mapscalefigsize\textwidth}
		\includegraphics[width=1.0\textwidth]{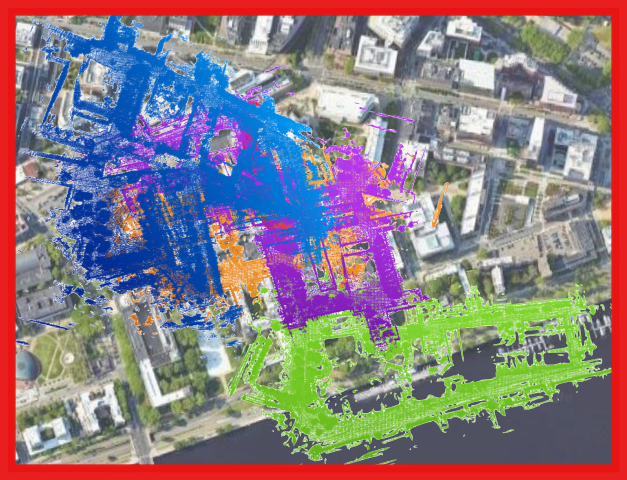}
		\caption{STD~\cite{Yuan23icra-STD}}
	\end{subfigure}
	\begin{subfigure}[b]{\mapscalefigsize\textwidth}
		\includegraphics[width=1.0\textwidth]{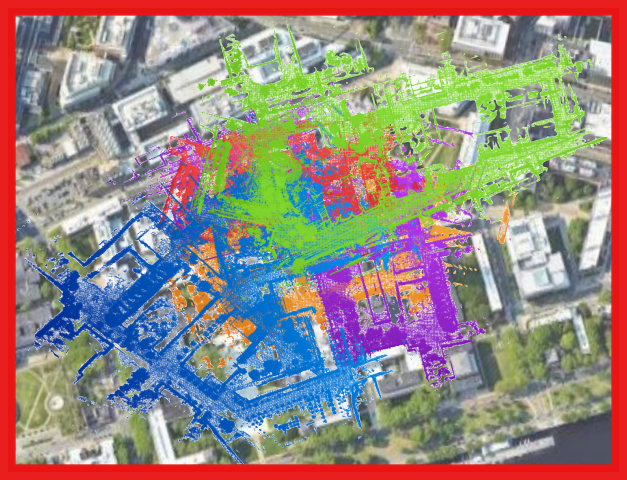}
		\caption{MapClosures~\cite{Gupta24icra-MapClosures}}
	\end{subfigure}
	\begin{subfigure}[b]{\mapscalefigsize\textwidth}
		\includegraphics[width=1.0\textwidth]{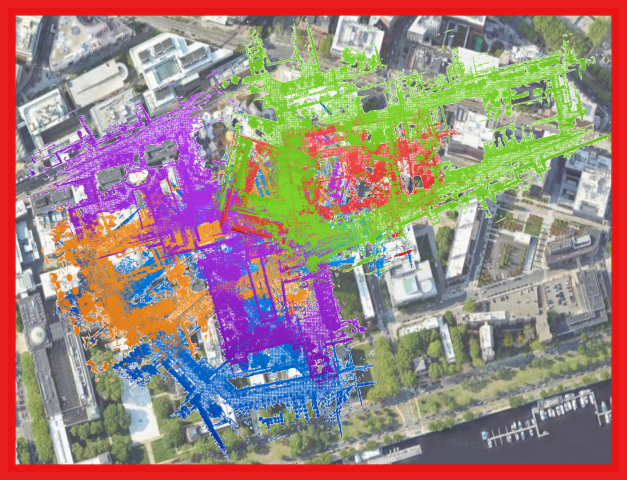}
		\caption{Predator~\cite{Huang21cvpr-PREDATORRegistration}}
	\end{subfigure}
	\begin{subfigure}[b]{\mapscalefigsize\textwidth}
		\includegraphics[width=1.0\textwidth]{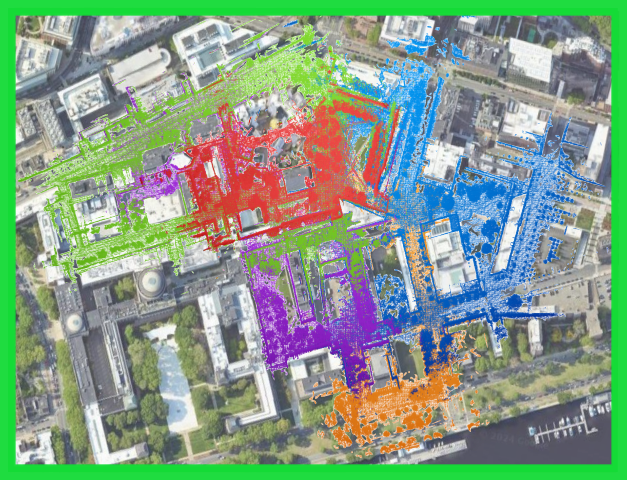}
		\caption{Proposed}
	\end{subfigure}
	\caption{(a)~Map clouds generated by FAST-LIO2-based SLAM~\cite{xu22tro-FastLIO2} in the Kimera-Multi dataset~\cite{Tian23iros-KimeraMultiExperiments}, with different colors representing different robots~(best viewed in color).
		(b)-(f)~Map-level registration comparison with the state-of-the-art approaches. Red and green boxes, outlining subfigures, indicate failure and success in registration, respectively. Note that the purple point cloud is the target cloud, so other point clouds should be aligned with the purple point cloud if the registration succeeds.}
	\label{fig:map_level_registration}
	\vsfig
\end{figure}

\subsection{Ablation Studies and Runtime Analyses} \label{sec:runtime}

\begin{figure}[t!]
	\captionsetup{font=footnotesize}
	\centering
	\begin{subfigure}[b]{0.48\textwidth}
		\includegraphics[width=1\textwidth]{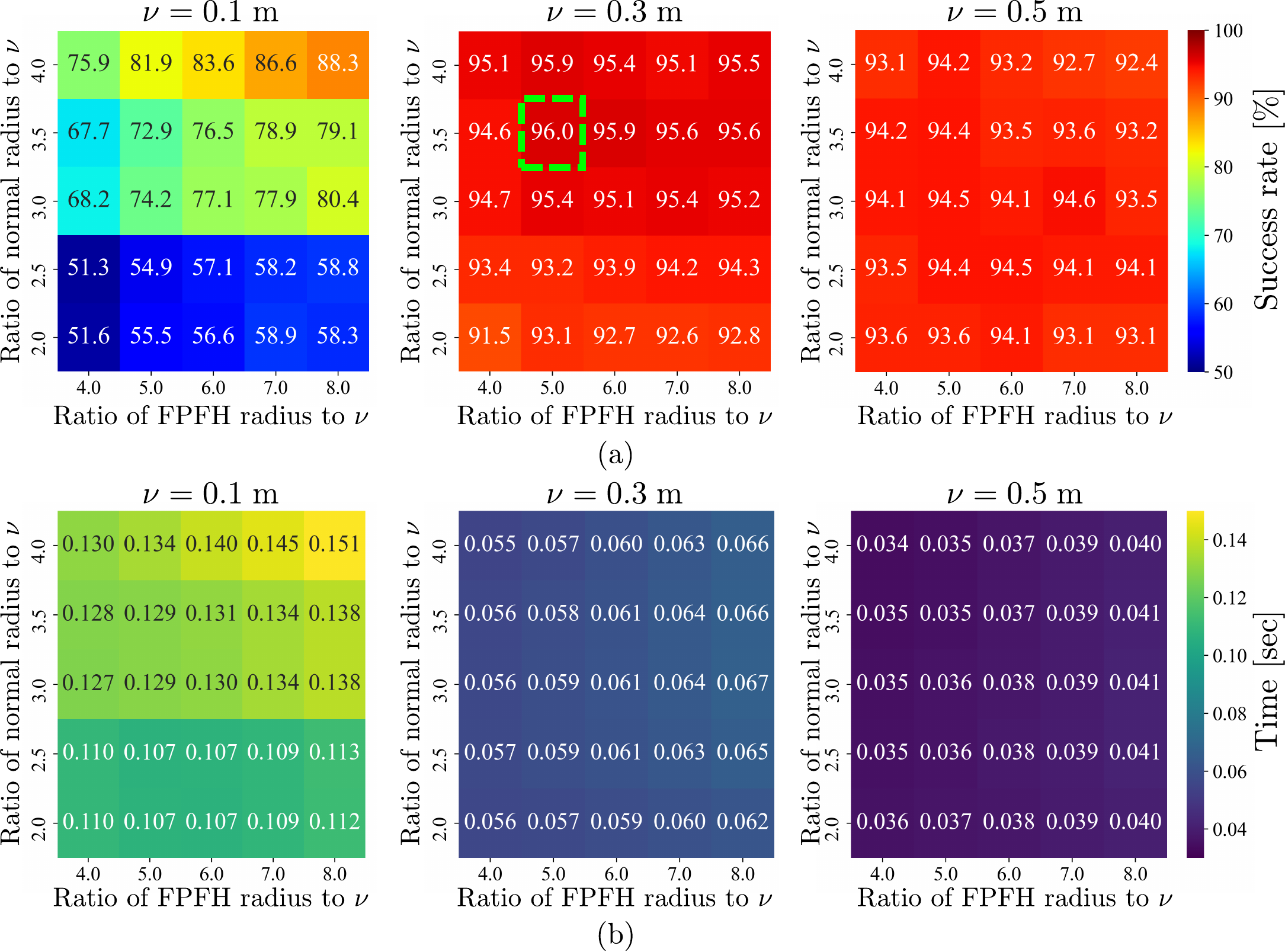}
	\end{subfigure}
	\caption{(a) Average success rates and (b) overall mean speed in the $10 \btwdist 12$\;m loop closing test using the KITTI and MulRan datasets for single scan-level registration, with varying values of normal radius,~$r_\text{normal}$, and FPFH radius,~$r_\text{FPFH}$ on an Intel(R) Core(TM) i9-13900 CPU. The green dashed box highlights the highest success rate~(best viewed in color).}
	\label{fig:fpfh_param_set}
\end{figure}

Finally, we conducted ablation studies and investigated the runtime of our KISS-Matcher.
As presented in \figref{fig:fpfh_param_set}, contrary to the common belief that the normal radius and FPFH radius should be set at 2 and 5 times the voxel size~\cite{Rusu09icra-fast3Dkeypoints,Rasu11icra,Zhou18arxiv-open3D}, respectively,
we showed that the optimal parameters are 3.5 and 5.0 times the voxel size, respectively, as they achieved the highest success rate while maintaining sufficiently fast processing speed.
This suggests that a larger normal radius is required to address the sparsity of point clouds from 3D LiDAR sensors.
Moreover, as presented in \figref{fig:mulran_comparison}(b), we also showed that setting $\tau_\text{lin}=0.99$ slightly enhanced the registration performance.

\begin{figure}[t!]
	\captionsetup{font=footnotesize}
	\centering
	\begin{subfigure}[b]{0.24\textwidth}
		\includegraphics[width=1.0\textwidth]{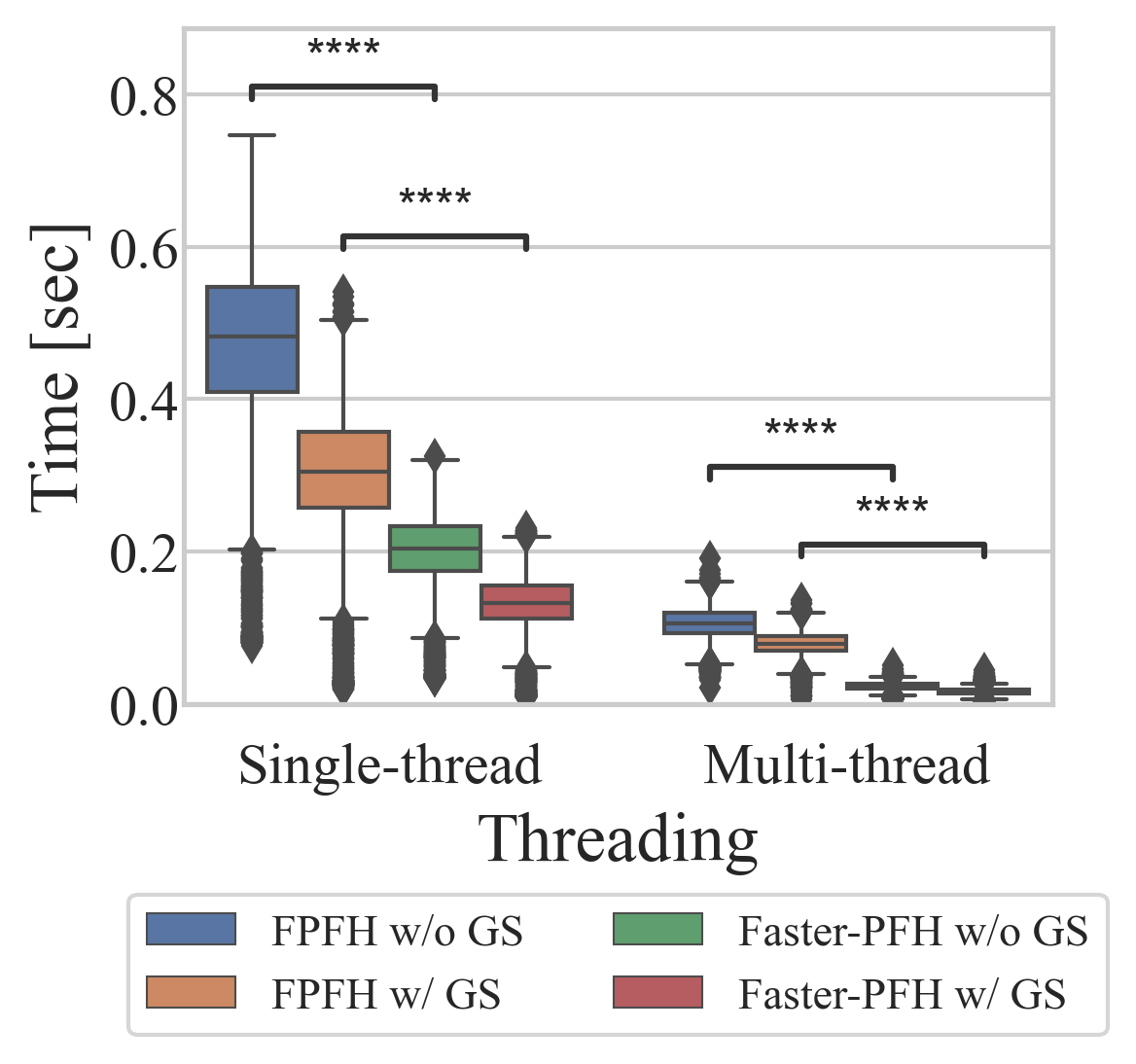}
	\end{subfigure}
	\begin{subfigure}[b]{0.235\textwidth}
		\includegraphics[width=1.0\textwidth]{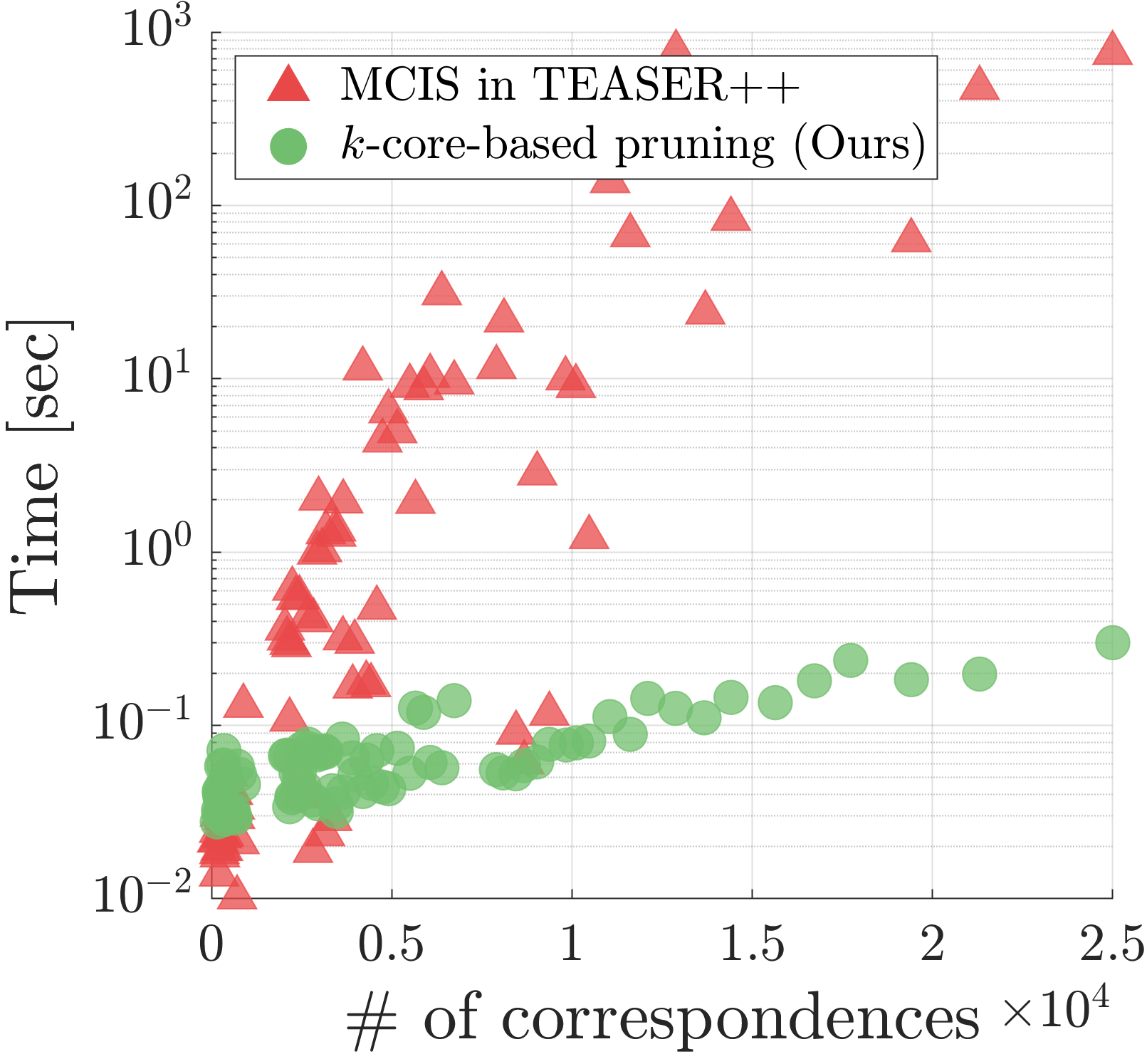}
	\end{subfigure}
	\caption{(a) Speed comparison between FPFH~\cite{Rusu09icra-fast3Dkeypoints} and our Faster-PFH under single-threading and multi-threading conditions in the $2 \btwdist 12$\;m loop closing test using the KITTI and MulRan datasets.
	The **** annotations indicate measurements with a $p$-value $< 10^{-4}$ after a paired $t$-test. Here, GS indicates geometric suppression in \figref{fig:overview} and we used ground segmentation~\cite{Lim21ral-Patchwork} as the method for geometric suppression.
	(b)~Speed comparison of the maximum clique inlier selection~(MCIS) in the TEASER++ library~\cite{Yang20tro-teaser} and our maximum $k$-core-based outlier rejection~\cite{Shi21icra-robin} given the same number of correspondences.}
	\label{fig:performance_comparison}
\end{figure}

Furthermore, we demonstrate that our Faster-PFH is much faster than the existing FPFH~(\figref{fig:performance_comparison}(a))
and our $k$-core-based graph-theoretic outlier rejection is much faster than the graph-theoretic module in TEASER++~\cite{Yang20tro-teaser}, especially when a large number of correspondences are given~(\figref{fig:performance_comparison}(b)).
Consequently, our approach can achieve a substantial speed-up with respect to TEASER++ in large-scale registration at the kilometer level, as shown in \figref{fig:main_fig}(b).
Notably, our KISS-Matcher exhibited more than a 20$\times$ speed improvement when the number of voxelized point clouds exceeded 200K.

%

%
\section{Conclusion}
\label{sec:conclusion}

In this study, we have revisited global point cloud registration from a holistic perspective and proposed a versatile open-source registration pipeline, \textit{KISS-Matcher}, designed to achieve fast, robust, and scalable registration from scan to map level.
Through experiments in various scenarios, we have demonstrated that our KISS-Matcher is on par with state-of-the-art approaches on the KITTI benchmark, while being much faster and exhibiting strong generalization across different datasets and scales, making it a practical solution for real-world applications.
In future works, we plan to apply our matching pipeline in mapping and localization applications.

\newif\ifisArXivMode

\isArXivModetrue

\ifisArXivMode
\section*{Acknowledgments}

We thank Prof. Cyrill Stachniss’ group at the University of Bonn, particularly Tizziano Guadagnino, Benedikt Mersch, Louis Wiesmann, and Jens Behley, for allowing us to use the term \textit{KISS}.
Especially, we thank Jens Behley for the fruitful discussions about tree architecture for neighbor search and Kenji Koide for open-sourcing the multi-threaded NanoFLANN tree~\cite{Koide24joss-smallgicp}, which we use in our Faster-PFH.
\else

\clearpage
\newpage

\fi

\bibliographystyle{IEEEtran}
\bibliography{../references/myRefs.bib,../references/refs.bib}

\ifisArXivMode
	\appendices
\renewcommand{\thesection}{A\arabic{section}}
\renewcommand{\theequation}{A\arabic{equation}}
\renewcommand{\thefigure}{A\arabic{figure}}
\renewcommand{\thetable}{A\arabic{table}}

\setcounter{equation}{0}
\setcounter{section}{0}
\setcounter{figure}{0}

\newcommand{\search}{\mathcal{S}}
\newcommand{\Pqvalid}{\mathcal{P}_\text{valid}}
\newcommand{\normalvec}{\mathbf{n}}
\newcommand{\plin}{p_\text{lin}}
\newcommand{\numthr}{\tau_\text{num}}
\newcommand{\neighboring}{k}

\newcommand{\neighboringIdxSet}{\mathcal{I}}
\newcommand{\neighboringValidIdxSet}{\mathcal{I}_\text{valid}}
\newcommand{\annot}[1]{\textcolor{gray!80}{\% #1}}

\newcommand{\searchfpfh}{\search_{\rfpfh}}
\newcommand{\validQueryIndices}{\mathcal{J}}

\section{Detailed Explanation of Faster-PFH}
\newcommand{\kd}{K-$d$}

To help the reader's understanding, we provide more details about our Faster-PFH, which is illustrated in \secref{subsec:fpfh}.
In particular, we describe the details of each dashed box from left to right in \figref{fig:fpfh_and_ours}(b).
The pseudocode corresponding to the leftmost dashed box is presented in Algorithm~\ref{alg:faster_pfh}.

First, let us define the radius neighbor search function using \kd~tree $\searchfpfh(\cdot, \cdot)$, which outputs the indices of neighboring points~$\mathcal{I}$, as follows:

\begin{equation}
 \label{eqn:radius_neighbor_search}
 \mathcal{I} = \searchfpfh(\querypoint, \mathcal{P})=\{s \mid\|\srcpoint_s-\querypoint\|<\rfpfh,\, \srcpoint_s \in \mathcal{P}\},
\end{equation}

\noindent where $\querypoint$ is a query point, $\rfpfh$ is the radius of the FPFH sphere, and $\mathcal{P}$~is a point cloud, which can be either the source or target cloud; $s$ indicates an index of a point in $\mathcal{P}$.
Then, by using \eqref{eqn:radius_neighbor_search}, we fetch the neighboring points $\Pfpfh$ that corresponds to the indices in $\neighboringIdxSet$~(\ie~``RS with $\rfpfh$'' in the leftmost dashed box of \figref{fig:fpfh_and_ours}(b)).

This radius search is intended to create a feature descriptor by utilizing the geometric properties between the query point and its neighboring points.
However, if the number of neighboring points around the query point $\querypoint$ is too small, the normal vector of $\querypoint$ is likely to be more imprecise, eventually degrading the quality of feature descriptors.
Therefore, if $|\Pfpfh| < \tau_\text{num}$, we skip the normal estimation and this query point is excluded from the following procedures~(line~\ref{line:skip}).
Otherwise, we subsample $\Pnormal$ from $\Pfpfh$ (\ie~``Subsamping with $\rnormal$'' in \figref{fig:fpfh_and_ours}(b)).

Next, if $|\Pnormal| \geq \tau_{\text{num}}$, the normal vector $\normalvec$ and its linearity $\plin=\frac{\lambda_{1}-\lambda_{2}}{\lambda_{1}}$ are estimated by PCA.
Note that if $\plin$ is too large, it indicates that $\Pnormal$ follows a line or pole-like distribution, resulting in an ill-defined normal vector.
This is because the orthogonal direction to the line is not uniquely determined.
For this reason, only if $\plin < \tau_\text{lin}$ is satisfied, $q$, $\normalvec$, and $\neighboringIdxSet$ are saved~(lines~\ref{line:cond_lin}--\ref{line:I}). Those are used when computing the feature descriptors.

During this process, there are indices that belong to $\neighboringIdxSet$ but are not assigned with a normal vector.
These indices typically correspond to the points located at the boundary of the 3D point cloud, where there are not enough points in their vicinity for normal estimation but these points are included in the neighboring searches of other points.
Thus, we reject these indices and update $\neighboringIdxSet$ as $\neighboringValidIdxSet$, where $\neighboringValidIdxSet \subseteq \neighboringIdxSet$.
Consequently, only the indices each of whose $\neighboringValidIdxSet$ satisfies $|\neighboringValidIdxSet| \geq \numthr$ are set as the final valid query index set $\validQueryIndices$; see lines~\ref{line:second_condition_start}--\ref{line:second_condition_end}.

\begin{algorithm}[t!]
{\footnotesize
\SetAlgoLined
\textbf{Input:} \ Voxelized point cloud $\srccloud$ with voxel size $v$; radiuses~$\rfpfh$ and $\rnormal$; user-defined thresholds $\numthr$ and $\tau_\text{lin}$. \\
\textbf{Output:} \ Valid query index set $\validQueryIndices $, normal vector set $\mathcal{V}$, and neighboring indices set $\mathcal{M}$ \\
Initialize K-$d$ tree for $\srccloud$  \\
    $N = |{\srccloud}|$ \\
    $\mathcal{B} = \texttt{allocate\_array}(N, \texttt{false})$ \annot{Boolean set indicating whether each index is reliable} \\
    $\validQueryIndices = \varnothing $ \annot{Index set each of whose corresponding normal vector is reliable} \\
    $\mathcal{V} = \texttt{allocate\_array}(N) $ \annot{Normal vector set} \\
    $\mathcal{M} = \texttt{allocate\_array}(N) $ \annot{Neighboring indices set} \\
\annot{Step 1. Calculate only reliable normal vectors} \\
\For{$q \leftarrow 1$ \normalfont{to} $N$ }{
    $\querypoint \in \srccloud$ \\
    $\neighboringIdxSet = \searchfpfh(\querypoint, \srccloud)$ \annot{$\searchfpfh(\cdot, \cdot):$ neighboring search} \\
    Fetch neighboring cloud points $\Pfpfh$ from $\srccloud$ using $\neighboringIdxSet$ \\
    \textbf{if} $ |{\Pfpfh}| < \numthr$ \textbf{then continue} \annot{Skip normal estimation} \label{line:skip}\\
    Subsample $\Pnormal$ from $\Pfpfh$ with radius $\rnormal$ \\
    \If{\normalfont $|\Pnormal| \geq \numthr$}{ \label{line:cardinality} \label{line:cond_num}
        $[\mathbf{n}, p_\text{lin}]$ = \texttt{EstimateNormalVector}($\Pnormal$) \\
        \If{\normalfont $p_\text{lin} < \tau_\text{lin}$}{ \label{line:cond_lin}
            $\validQueryIndices.\texttt{push\_pack}(q) $ \\
            $\mathcal{B}[q] = \texttt{true}$ \\
            $\mathcal{V}[q] = \mathbf{n} $ \\
            $\mathcal{M}[q] = \neighboringIdxSet$ \\ \label{line:I}
        }
    }
}

\annot{Step 2. Exclude indices identified as unreliable due to filtering in lines \ref{line:cond_num} and \ref{line:cond_lin}} \\
\For{$q$ {\normalfont in} $\validQueryIndices$}{ \label{line:second_condition_start}
    $\neighboringIdxSet_\text{valid} = \varnothing$ \\
    \For{$i$ {\normalfont in} $\mathcal{M}[q]$}{
        \annot{Set the indices whose normal vectors are assigned} \\
        \If{$\mathcal{B}[i]$}{
            $\neighboringIdxSet_\text{valid}.\texttt{push\_back}(i)$ \\
        }
    }
    $\mathcal{M}[q] \leftarrow \neighboringIdxSet_\text{valid}$ \\
}
\For{$q$ {\normalfont in} $\validQueryIndices$}{
    \If{\normalfont $\neighboringIdxSet_\text{valid} < \numthr$}{
        \annot{Remove the query index that has only few valid neighboring indices} \\
        $\mathcal{B}[q] = \texttt{false}$ \\
        Remove $q$ from $\validQueryIndices$ \\
    }
} \label{line:second_condition_end}
}
    \caption{Neighboring search, normal estimation, and linearity-based filtering in Faster-PFH\label{alg:faster_pfh}}
\end{algorithm}

Finally, we compute the simplified point feature histogram~(SPFH) and fast point feature histograms~(FPFH) feature for the reliable points~(\ie~indices in $\validQueryIndices$).
Let the neighboring point of the query point $\querypoint$ be denoted as~$\srcpoint_\neighboring$, where $\neighboring \in \neighboringValidIdxSet$.
Among these two points, the one with the smaller angle between the vector connecting the points,~\ie~$\srcpoint_k-\querypoint$, and its estimated normal is designated as $\srcpoint_\pfhi$, while the one with the larger angle is selected as $\srcpoint_\pfhj$; see the yellow angles in \figref{fig:angular_feature}.

Next, let the corresponding normal vectors of $\srcpoint_\pfhi$ and $\srcpoint_\pfhj$ be denoted by $\mathbf{n}_\pfhi$ and $\mathbf{n}_\pfhj$, respectively, and the direction vector between $\srcpoint_\pfhi$ and $\srcpoint_\pfhj$ be defined as $\boldsymbol{d}=\frac{\srcpoint_\pfhj - \srcpoint_\pfhi}{\twonorm{\srcpoint_\pfhj - \srcpoint_\pfhi}}$, where $\twonorm{\cdot}$ denotes the $L2$-norm.
Based on the Darboux~$\boldsymbol{u}\boldsymbol{v}\boldsymbol{w}$ frame~\cite{Rusu09icra-fast3Dkeypoints}, three angular features $\text{f}_1$, $\text{f}_2$, and $\text{f}_3$ are defined as follows:

\begin{equation}
 \label{eqn:angular_features}
 \begin{aligned}
  \text{f}_1 & ={\rm atan2}\left(\boldsymbol{w} \cdot \mathbf{n}_\pfhj, \boldsymbol{u} \cdot \mathbf{n}_\pfhj\right), \; \text{f}_2 =\boldsymbol{v} \cdot \mathbf{n}_\pfhj, \; \text{f}_3 =\boldsymbol{u} \cdot \boldsymbol{d},
 \end{aligned}
\end{equation}

\noindent where $\boldsymbol{u} = \mathbf{n}_\pfhi$, $\boldsymbol{v} = \boldsymbol{d} \times \boldsymbol{u}$, $\boldsymbol{w} = \boldsymbol{u} \times \boldsymbol{v}$, $\cdot$ denotes the inner product, $\times$ denotes the cross product, and ${\rm atan2}(y, x)= \arctan(\frac{y}{x})$.

Next, let us define a mapping function $g_\histidx(\cdot, \cdot)$ that maps the $l$-th angular feature $\text{f}_l$ where $l \in \{1, 2, 3\}$ to the index of the $l$-th histogram as follows:
\begin{equation}
 \label{eqn:histogram}
 g_\histidx(\querypoint, \srcpoint_k) = \lfloor H \frac{\text{f}_{\histidx}-\text{f}_{\histidx, \min}}{\text{f}_{\histidx, \max} + \epsilon - \text{f}_{\histidx, \min}}\rfloor + 1,
\end{equation}

\noindent where $\lfloor \cdot \rfloor$ is the floor function, $H$ denotes the size of a histogram, and $\epsilon > 0$ is a small positive number to prevent $g_\histidx(\querypoint, \srcpoint_k)$ from incorrectly returning $H+1$ when $\text{f}_\histidx$ equals to $\text{f}_{\histidx, \max}$; $\text{f}_{\histidx, \min}$ and $\text{f}_{\histidx, \max}$ represent the minimum and maximum values of $\text{f}_l$, respectively.
Then, the SPFH signature of $\srcpoint_q$ is defined as $\text{SPFH}(\srcpoint_q) = [\mathbf{f}_1 \; \mathbf{f}_2 \; \mathbf{f}_3] \in \mathbb{R}^{3H}$, each of whose $h$-th element $\mathbf{f}_l[h]$ is defined as follows:

\begin{equation}
 \label{eqn:each_of_spfh}
 \mathbf{f}_l[h]=\frac{100.0}{|\neighboringValidIdxSet|} \sum_{\neighboring \in \neighboringValidIdxSet} \llbracket g_\histidx\left(\querypoint, \srcpoint_k\right)=h \rrbracket,
\end{equation}

\noindent where $\llbracket \cdot \rrbracket$ outputs one if the condition is satisfied and zero, otherwise.

Finally, the FPFH feature of $\querypoint$, $\text{FPFH}(\srcpoint_q)$, is defined as follows:

\begin{equation}
 \label{eqn:fpfh}
 \text{FPFH}(\querypoint)=\text{SPFH}(\querypoint)+\frac{1}{|\neighboringValidIdxSet|} \sum_{\neighboring \in \neighboringValidIdxSet} \frac{1}{\omega_k} \text{SPFH}\left(\srcpoint_k\right),
\end{equation}

\noindent where $\omega_k$ represents the distance between $\querypoint$ and $\srcpoint_k$.

In summary, we do not propose a new type of descriptor, but we significantly improve the speed by reducing redundant parts as much as possible and filtering out unreliable points in advance.
In addition, as you can see, an unreliable normal vector makes the angular features in \eqref{eqn:angular_features} imprecise, which in turn negatively affects the quality of SPFH and FPFH calculations, from \eqref{eqn:histogram} to \eqref{eqn:fpfh}.
Thus, this also supports our rationale for discarding points with unreliable normal vectors.

\begin{figure}[t!]
    \captionsetup{font=footnotesize}
    \centering
    \begin{subfigure}[b]{0.35\textwidth}
        \includegraphics[width=1.0\textwidth]{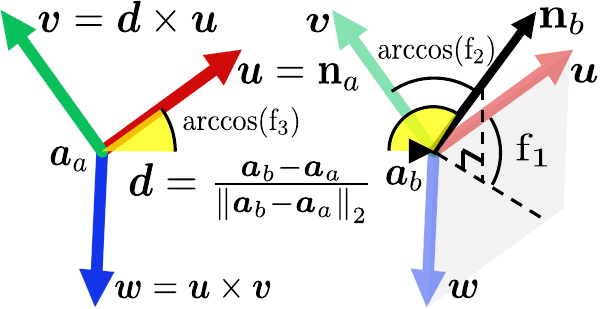}
    \end{subfigure}
    \caption{Visual description of angular features~$\text{f}_1$, $\text{f}_2$, and $\text{f}_3$. Note that $\text{f}_1$ geometrically means the angle between the projection of $\mathbf{n}_\pfhj$ onto the $\boldsymbol{u}\boldsymbol{w}$-plane and $\boldsymbol{u}$.}
    \label{fig:angular_feature}
\end{figure}

\fi

\end{document}